\documentclass{iosart2c}

\usepackage[T1]{fontenc}
\usepackage{times}%
\pdfoutput=1
\usepackage{amsmath}
\usepackage{dcolumn}
\usepackage{graphicx}

\usepackage{algorithmic}
\usepackage{amsmath, amsfonts,amssymb,amsthm}
\usepackage{multirow}
\usepackage{subfigure}

\newcolumntype{d}[1]{D{.}{.}{#1}}

\firstpage{1} \lastpage{5} \volume{1} \pubyear{2009}

\begin{document}
\begin{frontmatter}                           

%
\title{Weight-based Fish School Search algorithm for Many-Objective Optimization}

\runningtitle{Weight-based Fish School Search algorithm for Many-Objective Optimization}

\author[A]{\fnms{Fernando} \snm{Buarque de Lima Neto}\thanks{Corresponding author. E-mail: fbln@ecomp.poli.br.}},
\author[A]{\fnms{Isabela Maria} \snm{Carneiro de Albuquerque}}
and
\author[A]{\fnms{Jo\~ao Batista} \snm{Monteiro Filho}}
\runningauthor{F. B. de Lima Neto et al.}
\address[A]{Department of Computer Engineering, Polytechnical School of Pernambuco, Rua Benfica, 455, Recife, Brazil\\
E-mail: \{fbln,imca,jbmf\}@ecomp.poli.br}

\begin{abstract}
Optimization problems with more than one objective consist in a very attractive topic for researchers due to its applicability in real-world situations. Over the years, the research effort in the Computational Intelligence field resulted in algorithms able to achieve good results by solving problems with more than one conflicting objective. However, these techniques do not exhibit the same performance as the number of objectives increases and become greater than 3. This paper proposes an adaptation of the metaheuristic Fish School Search to solve optimization problems with many objectives. This adaptation is based on the division of the candidate solutions in clusters that are specialized in solving a single-objective problem generated by the decomposition of the original problem. For this, we used concepts and ideas often employed by state-of-the-art algorithms, namely: (i) reference points and lines in the objectives space; (ii) clustering process; and (iii) the decomposition technique Penalty-based Boundary Intersection. The proposed algorithm was compared with two state-of-the-art bio-inspired algorithms. Moreover, a version of the proposed technique tailored to solve multi-modal problems was also presented. The experiments executed have shown that the performance obtained by both versions is competitive with state-of-the-art results.
\end{abstract}


\end{frontmatter}

\section{Introduction}
Real-world problems often involve optimizing more than one objective. In many cases, these objectives are in conflict with each other and there might not be a single solution capable of simultaneously optimizing all goals. These problems are known as multi-objective optimization problems (MaOP). In order to solve them, candidate algorithms should provide as answers a set of good ``trade-off" solutions so that a decision maker can pick the most suitable ones \cite{reyes2006multi}. More precisely, a multi-objective optimization can be defined by means as Pareto-dominance \cite{deb2001multi}.

Not rarely, in real-world problems many (i.e. more than 3) objectives need to be considered \cite{deb2014evolutionary}. In those cases, solving the optimization problem is even more difficult due to the fact the increase on the number of objectives makes harder to distinguish good solution from the others \cite{yuan2016new}. MaOPs appear in several domains such as software engineering \cite{praditwong2011software} and automotive engine calibration \cite{lygoe2013real}. 
 
Research on evolutionary and swarm intelligence algorithms for solving MaOPs features straightforward applications of existing multi-objective algorithms for problems with 2 or 3 objectives. Despite their success in solving problems with 2 objectives, these algorithms face difficulties in solving many-objective optimization problems. The most important challenge faced by multi-objective algorithms in this scenario, mostly for the Pareto-based ones \cite{deb2001multi}, is the search ability deterioration with the increase on the number of objectives. When this quantity grows, almost all solutions in the population become non-dominated, i.e. equally good \cite{ishibuchi2008evolutionary}. In those cases, evolutionary algorithms lose their selection pressure and swarm intelligence algorithms lose their exploitation ability. As a consequence, algorithms' capability of converging to good solutions is severely deteriorated. In other words, multi-objective algorithms scale up poorly in many-objective optimization problems \cite{castro2016c}. In order to tackle this issue, researchers in the field have proposed different solutions, such as adopting new preference relations as fuzzy Pareto-dominance \cite{farina2004fuzzy} and $\epsilon$-dominance \cite{laumanns2002combining}.

Other approach used by researchers to deal with the loss of convergence capability is the decomposition of the MaOP in simpler single-objective problems, as it is done in the MOEA/D \cite{zhang2007moea}. Algorithms based on this idea have shown good performance \cite{yuan2015multiobjective}. Another advantageous practice in the research of bio-inspired algorithms for solving MaOPs is the use of reference points in the objectives space. The goal is to guide the search process to enforce diversity among the solutions returned by the algorithm. This is an important factor to be considered because many algorithms tend to favor dominance resistant solutions \cite{ikeda2001failure}, which are solutions with high performance in at least one of the objectives, but with especially poor performance in the other. An example of successful use of reference points is the NSGA-III algorithm \cite{deb2014evolutionary}. In recent years some researchers are dedicating their attention to combine both decomposition-based approach and reference points for enhancing the performance of many-objective optimization algorithms. An example of state-of-the-art method that uses both techniques is the Many-objective Evolutionary Algorithm Based on Dominance and Decomposition (MOEA/DD) \cite{li2015evolutionary} and the $\theta$-Dominance Evolutionary Algorithm ($\theta$-DEA) \cite{yuan2016new}.

This work incorporates the decomposition-based approach and the use reference points to the swarm intelligence algorithm Fish School Search (FSS) \cite{bastos2008novel} with the goal of enabling it to solve MaOPs. FSS versions produced so far already presented good performance in optimizing multi-modal functions \cite{buarque2014weight} (\textit{e.g.} multi-plateau functions \cite{monteiro2016}) and multi-objective problems \cite{bastos2015multi}. Also, when compared to other swarm intelligence algorithms, one of FSS's control mechanisms, the weights of the fishes, natively provides a way to store how good the solutions found by the algorithm are. This is a relevant aspect as most of other swarm intelligence techniques proposed to solve MaOPs use an external archive to store good solutions to guide the search process. The methods chosen to prune, update and select solutions from the archive are crucial to algorithms achieve good performances \cite{figueiredo2016many}. Therefore, the schemes applied to manage such archives are drawing considerable attention from researchers in the field \cite{reyes2006multi}.

The central idea of this work is to adapt a multi-modal version of FSS, the weight-based Fish School Search (wFSS) \cite{buarque2014weight}, to solve problems with many objectives without the need of storing the best solutions found throughout the search in an external archive. The proposed technique, named weight-based many-objective Fish School Search (wmoFSS), has the same operators as the original version of FSS and a new clustering operator is included to promote diversity during the searching process. This clustering process is responsible to split the swarm in sub-groups. Moreover, in wmoFSS the MaOP is decomposed in sub-problems and each one is assigned to one sub-group that tries to solve it. Hence, the whole swarm solves all sub-problems at the same time. In comparison with the original version of FSS, the most important difference in wmoFSS is the fact that each fish has now a vector $\mathbf{w}$ of weights in which a single component represents the correspondent weight of a fish considering the objectives individually. Furthermore, a
modified version of the proposed technique was also proposed with the goal of improving its performance on multi-modal many-objective problems.

This paper is organized as follows: Section \ref{Prelim} describes the main concepts related to wmoFSS; Section \ref{wmoFSS} presents the novel algorithm; In Section \ref{Experiments} we describe the experimental design, the results and analysis; In Section \ref{SBX}, we propose a modified version of wmoFSS for solving multi-modal problems and present further experiments; In Section \ref{Conclusion} we draw conclusions related to this work and point out potential future works.

\section{Background} \label{Prelim}
\subsection{Fish School Search algorithm}
The Fish School Search (FSS) is a population-based metaheuristic algorithm inspired in the behavior of swimming fishes that expand and contract while looking for food. Each fish $n$-dimensional location represents a possible solution for the optimization problem. Each solution cumulative account on how successful has been the search is represented by a feature named weight. 

FSS is composed by the feeding and movement operators, the latter being split into three components: individual, collective-instinctive and collective-volitive. The individual component of the movement allows every fish in the school to perform a random local search looking for promising regions in the search space. This component is computed according to the following equation: 
\begin{equation}\label{IndMovFSS}
	\textbf{x}_i(t+1)=\textbf{x}_{i}(t)+\mathbf{r}\text{step}_{\text{ind}},
\end{equation}
where $\textbf{x}_{i}(t)$ and $\textbf{x}_{i}(t+1)$ represent the position of a fish $i$ before and after the movement caused by the individual component, respectively. $\mathbf{r}\in\mathbb{R}^N$ with $ r_j \sim \text{Uniform}[-1, 1]$, for $j=\{1, \ldots, n\}$. $\text{step}_{\text{ind}}$ is a parameter responsible to set the maximum displacement for this movement. A new position $\textbf{x}_i(t+1)$ is only accepted if (for maximization) $f(\textbf{x}_i(t+1)) > f(\textbf{x}_i(t))$, where $f$ is the objective function. Otherwise, the fish remains in the same position and $\textbf{x}_i(t+1)=\textbf{x}_{i}(t)$.

The collective-instinctive component of the movement is the average of individual movements for all $\textbf{x}_i$. A vector $\textbf{I}\in\mathbb{R}^N$ representing the weighted average of displacements for each $\textbf{x}_i$ is calculated according to:
\begin{equation}\label{IVec}
	\textbf{I}=\frac{\sum^{S}_{i=1} \Delta \textbf{x}_{i} \Delta f_{i}}{\sum^{N}_{i=1} \Delta f_{i}},
\end{equation}
where $S$ is the size of the school, $\Delta \textbf{x}_{i}$ is a shorthand for $\textbf{x}_i(t+1) - \textbf{x}_i(t)$, and $\Delta f_i$ is a shorthand for $f(\textbf{x}_i(t+1)) - f(\textbf{x}_i(t))$. The displacement represented by $\textbf{I}$ is defined in a way that fishes with a higher improvement will attract other fishes to its position. After computing $\textbf{I}$, every fish moves according to:
\begin{equation}
	\textbf{x}_i(t+1)=\textbf{x}_{i}(t)+\textbf{I}.
\end{equation}

The collective-volitive component is used in order to regulate school's exploration/exploitation ability during the search process. First, the barycenter $\textbf{B}\in \mathbb{R}^N$ of the school is calculated based on each fish position $\textbf{x}_{i}$ and weight $w_{i}$, as described in Eq. \ref{barycenter}:
\begin{equation}\label{barycenter}
	\textbf{B}(t)=\frac{\sum^{S}_{i=1} \textbf{x}_{i}(t) w_{i}(t)}{\sum^{N}_{i=1} w_{i}(t)}.
\end{equation}
Then, if the total school weight $\sum^{S}_{i=1} w_{i}$ has increased from $t$ to $t+1$, fishes move towards $\textbf{B}$. Otherwise, fishes are spread away from the barycenter. 

Besides the movement operators, the feeding operator is responsible for updating the weights according to:
\begin{equation}\label{Feeding}
	w_{i}(t+1)=w_{i}(t)+\frac{\Delta f_i}{\text{max}(| \Delta f_i |)},
\end{equation}
$w_i(t)$ is only allowed to vary from 1 up to $w_{\text{scale}}$, which is a hyper-parameter. All weights are initialized with the value $w_{\text{scale}}/2$.

\subsection{Weight-based Fish School Search Algorithm}
Introduced in the work of Lima-Neto and Lacerda \cite{buarque2014weight}, wFSS is a weight-based niching version of FSS intended to provide multiple solutions in a single run for multi-modal optimization problems. The niching strategy is based on a new operator called Link Formation. This operator is responsible for defining leaders for the fishes in order to form sub-swarms and works according to the following: each fish $a$ chooses randomly another fish $b$ in the school. If $b$ is heavier than $a$, then $a$ now has a link with $b$ and follows $b$ (\textit{i.e.} $b$ leads $a$). Otherwise, nothing happens. However, if $a$ already has a leader $c$ and the sum of weights of $a$ followers is higher than $b$ weight, then $a$ stops following $c$ and starts to follow $b$. In each iteration, if $a$ becomes heavier than its leader, the link will be broken. 

In addition to Link Formation operator inclusion, some modifications were performed in the computation of the collective components of the movement operators in order to emphasize leaders influence on each sub-swarm. 
Also, the Collective-volitive component of the movement is also modified in a sense that the barycenter is now calculated for each fish with relation to its leader. If the fish does not have a leader, its barycenter will be its current position. 

\section{Weight-based Many-Objective Fish School Search Algorithm}\label{wmoFSS}
The wmoFSS algorithm incorporates a series of modifications in FSS and some of its previously proposed versions (FSS-SAR \cite{monteiro2016} and FSS-NF \cite{monteiro2016improved}) in order to adapt it to solve many-objective optimization problems. In this section we describe in details the framework of wmoFSS and highlight the differences between this algorithm and the original version of FSS.

\subsection{Core Idea}
The main idea of wmoFSS is to split up the whole swarm into sub-swarms, as in wFSS, and also to decompose the MaOP into scalar sub-problems, as in MOEA/D. Unlike wFSS, wmoFSS substitutes the Link Formation operator by a Clustering operator similar to the proposed by Deb and Jain \cite{deb2014evolutionary}. Each sub-swarm is responsible to solve one sub-problem and all sub-swarms work simultaneously. Thus, the algorithm is able to provide multiple solutions in a single run. From the other versions of FSS, wmoFSS uses the Stagnation Avoidance Routine \cite{monteiro2016} in the Individual component of the movement and a Feeding operator similar to the one used in NF versions \cite{monteiro2016improved}. The wmoFSS algorithm also incorporates the reference points generation scheme and normalization procedure used in NSGA-III and a comparison criterion based on the $\theta$-dominance utilized in $\theta$-DEA algorithm.

In comparison with original FSS and all of its versions taken into account in this work, a difference wmoFSS comprises is the fact that the weight of a fish is represented by a vector $\textbf{w}$, instead of a single scalar value. Each component of $\textbf{w}$ represents the correspondent weight of a fish considering one objective individually. For a problem with $m$ objectives, the weight vector of a fish $i$ is $m$-dimensional and represented as:
\begin{equation}
\textbf{w}_i = [w_{i,1}, w_{i,2}, ..., w_{i,m}]^T.
\end{equation}
Another critical modification incorporated by wmoFSS is that the Feeding operator is also responsible for to aggregate the weight vector into a single scalar value named aggregated weight and denoted as $\overline{w}$. This values represents a measure of the quality of a fish and the $\theta$-dominance criterion takes it into account in order to compare fishes. The complete description of the aggregation method used to compute $\overline{w}$ is further explained in this section. The following pseudocode describes the main framework of wmoFSS. 
	\begin{algorithmic}[1]
		\STATE Create reference points; 
		\STATE Initialize fishes according to $\text{Uniform}[-1, 1]$;
		\STATE Run Clustering operator; 
		\WHILE{Stopping condition is not met} 
		\FOR{Each fish on the school} 
		\STATE Move each fish according to its Individual component;
		\STATE Run Feeding operator; 
		\STATE Leaders definition; 
		\STATE Move each fish according to its Collective-instinctive component; 
		\STATE Move each fish according to its Collective-volitive component; 
		\ENDFOR
		\ENDWHILE
		\STATE Sort each cluster based on Pareto-dominance; 
		\STATE Return non-dominated solutions from each cluster.
	\end{algorithmic}
	\label{wmoFSSpseudocode}
First, a set of reference points in the objectives space is generated, then the swarm is randomly initialized following an uniform distribution on the decision space and split in clusters. We used the two-layer reference points generation method proposed by Deb and Jain \cite{deb2014evolutionary}. Then, until a user-defined stopping criterion is not met, operators are applied in a sense that firstly fishes perform a greedy local search, then the Feeding operator is applied to each fish and the collective components of the movement are calculated  considering each cluster individually. At last, the algorithm sort each cluster based on the Pareto-dominance criterion \cite{deb2001multi} and return the non-dominated solutions of each cluster.

\subsection{Clustering Operator}
The Clustering operator of wmoFSS is responsible for splitting the population into sub-swarms or clusters, as well as the Link Formation of wFSS. This operator divides the school into clusters according to the perpendicular Euclidean distance between a fish and a set of reference lines in the objective space. A reference line $\boldsymbol\lambda_k$ is defined as a line that goes from the ideal point to a reference point generated by the previously defined approach. The assignment process executed by the Clustering operator is similar to the one used by NSGA-III algorithm. Considering a set of $N$ reference points and a set of $S$ candidate solutions, the $\lfloor \frac{S}{N'} \rfloor$ closest solutions to a reference line are assigned to it. If $S$ is not a multiple of $N'$, the $S(\text{mod}\;N')$ remaining fishes are assigned to its closest reference lines. Within this clustering procedure, we want to avoid the possibility that a reference line has no solution associated with and some regions of the PF might not be reached.

\subsection{Individual Component of the Movement}
As FSS and wFSS, the Individual component of the movement in wmoFSS is responsible for each fish greedy local search, as described by Equation \ref{IndMovFSS}. On the other hand, in wmoFSS the aggregated weight (further defined is this Section) is the attribute considered in the comparison between the current and the candidate positions. 

Furthermore, wmoFSS incorporates the Stagnation Avoidance Routine \cite{monteiro2016improved}. Hence, if the aggregated weight of the candidate position is smaller than the current one, the fish moves towards the new position. Otherwise, a random number is drawn from an uniform $\text{Uniform}[0,1]$ and compared with a hyperparameter $\alpha$ which decays along with the iterations. If the random number is smaller than $\alpha$, the fish moves to the candidate position, otherwise, this positions remains the same. 

\subsection{Feeding Operator}
In wmoFSS the weight of a fish is a vector with components corresponding to the success with respect to each objective. The Feeding operator is then responsible to calculate these components and also to aggregate them to provide a single score to each solution in the swarm.

Its first task is to normalize the solutions in the objective space. This step is necessary due to the fact the objectives might not vary in the same range and we want to avoid giving preference to one objective in spite of the others during the aggregation procedure. We decided to consider an adaptive normalization process as the extreme objectives values found by fishes vary within the search. 

Our normalization process is straightforward. All components of the objective vector of each fish are normalized into an interval which extremes are the minimum and maximum values found so far for this objective. It is noteworthy to highlight that the vector whose coordinates are the best values for the $m$ objectives is the ideal point and represented as $\textbf{z}^*$. If the ideal point of the problem is known, we use it. The vector which coordinates are the worst values found for each objective is an approximation of the nadir point (nadir objective vector), denoted as $\textbf{z}^{nad}$ \label{znad}. This vector is constructed from the worst values of each objective considering just solutions on the Pareto front (PF) \cite{deb2001multi}. However, we use an approximation of $\textbf{z}^{nad}$ due to its difficult estimation \cite{deb2010toward}. 

For a fish $i$ with objective vector $\textbf{f}_i(t)$, the $j$-th component of its weight vector, corresponding to the $j$-th objective function, is computed according to the following equation:         
\begin{equation}
w_{i,j}(t+1) = \frac{f_{i,j}(t) - z^*_j(t)}{f_{j, max}(t) - z^*_j(t)}.
\label{feedingRN}	
\end{equation}
$f_{j, max}(t)$ is the maximum value found for the $j$-th objective until iteration $t$ and $z^*_j$ is the $j$-th coordinate of the known ideal point or the current value found by the algorithm so far. 



After the calculation of the weight vector for all fishes in the school, the Feeding operator computes the aggregated weight $\overline{w}$ \label{aggregated} of each fish. In order to do this, we define that the aggregated weight of a fish $i$, with weight vector $\textbf{w}_i$, belonging to the cluster associated with the reference line $\lambda_k$, is the value of the achievement scalaring function (ASF) Penalty-based Boundary Intersection (PBI) obtained for this fish, which is calculated by the following equation \cite{zhang2007moea}:

\begin{equation}
\overline{w}_i = g^{pbi}(\textbf{w}_i| \boldsymbol\lambda_k) = d_{ 1}(\textbf{w}_i) + \theta d_{2}(\textbf{w}_i),
\end{equation}
where $d_{1}(\mathbf{w}_i)$ and $d_{2}(\textbf{w}_i)$ are given by:
\begin{equation}
d_{1}(\textbf{w}_i) = \frac{|| \textbf{w}_i^T \boldsymbol\lambda_k||}{||\boldsymbol\lambda_k||},
\end{equation}
\begin{equation}
d_{2}(\textbf{w}_i) = || \textbf{w}_i - d_{1}(\textbf{w}_i) \frac{\boldsymbol\lambda_k}{||\boldsymbol\lambda_k||} ||.
\end{equation}
$\theta$ is a user-defined parameter and here we chose $\theta = 5$ \cite{li2015evolutionary}. As we mentioned before, $d_{k, 2}$ consists in the Euclidean distance between $\textbf{w}_i$ and its orthogonal projection onto $\boldsymbol\lambda_k$ and represents a measure of how far $\textbf{w}_i$ is from the reference line associated with its respective cluster. $d_{1}$ is the Euclidean distance between the orthogonal projection of $\textbf{w}_i$ onto $\boldsymbol\lambda_k$ and the ideal point. 

It is important to highlight here that an ASF can be seen as a way of represent the ``success" obtained by a solution in a determined problem and there many ways to define those functions. Among all possible choices for achievement scalarization function, as the often used Weighted-sum and the Tchebycheff, we selected the PBI due to the fact this ASF have shown to provide good results when used to solve problem from the DTLZ test suite \cite{deb2014evolutionary}, the set of problems used to evaluate wmoFSS in this work. Furthermore, this function allows the user to control the balance between convergence and diversity by changing the value of $\theta$.  

\subsection{Leaders Definition}
After splitting the population into clusters, to feed the fishes and obtain the value of their aggregated weights, the leader of each cluster is defined. We used a criterion inspired on the $\theta$-dominance \cite{yuan2015multiobjective} to assign leadership in each sub-swarm. This dominance relation just considers individuals from the same cluster and it is based on the PBI aggregation method and consider as quality measure the value of the PBI achievement scalarizing function of a normalized objective vector in all reference directions considered by the algorithm. However, for wmoFSS, we want that each sub-swarm is specialized in solving its respective subproblem, thus we choose to consider just the reference direction associated with the cluster of a fish. Hence, according to this modified $\theta$-dominance, which we denoted as $\theta^*$-dominance, a fish $i$ dominates $(\prec_{\theta^*})$ a fish $j$ if and only if: (i) $i$ and $j$ belong to the same cluster; (ii) $\overline{w}_i$ $<$ $\overline{w}_j$.  

Based on the aforementioned definitions, the leaders assignment process consists in to sort each cluster according to $\theta^*$-dominance and assign leadership to all $\theta^*$-non-dominated solutions. Information regarding clusters, leaders and values of aggregated weight vectors is used to compute the collective components of the movement.

\subsection{Collective-Instinctive Component of the Movement Operator}
The Collective-instinctive component of the movement in wmoFSS is computed considering just fishes in the same cluster. The values of aggregated weight are used to measure the increase of success achieved for each fish, as in FSS-NF versions. This measure is defined as $\Delta \overline{w} = \overline{w}(t) - \overline{w}(t+1)$ for minimization problems and $\Delta \overline{w} = \overline{w}(t+1) - \overline{w}(t)$ for maximization problems. For a fish $i$, its position after the collective-instinctive movement is computed according to:  
\begin{equation}
\textbf{x}_i(t+1) = \textbf{x}_i(t) + L \left( \frac{\sum_{k \in \Phi_i} \Delta \textbf{x}_k \Delta		\overline{w}_k}{\sum_{k \in \Phi_i} \Delta\overline{w}_k} \right),
\end{equation}
where $\Phi_i$ is the cluster of fish $i$. The parameter $L$ is equal to 1 if $i$ is a cluster leader and 0 otherwise. As for the FSS-SAR versions, only the fishes which improved their aggregated weight are allowed to contribute to this component computation. 

\subsection{Collective-Volitive Component of the Movement Operator}
As in Collective-instinctive component computation, in the Collective-volitive component of the movement only fishes in the same cluster are considered for to determine the value of the barycenter, that is, instead of calculate school's barycenter, in wmoFSS a barycenter for each cluster is computed. Given a fish $i$, the barycenter of its cluster is computed as follows:   
\begin{equation}
\textbf{B}_i(t) =  \frac{\sum_{k \in \Phi_i} \textbf{x}_k \frac{1}{\overline{w}_k}}{\sum_{k \in \Phi_i} \frac{1}{\overline{w}_k}}. 
\end{equation}
The movement towards barycenter is computed exactly as in FSS. Clusters leaders are not affected by this operator as well as for the Collective-instinctive component. 

\section{Experimental Design} \label{Experiments}
To evaluate wmoFSS performance, we selected the first four problems of the DTLZ \cite{deb2002scalable} due to the fact they appear very often in literature \cite{von2014survey}. We compared the Inverted Generational Distance (IGD) metric \cite{bosman2003balance} obtained by wmoFSS with the results of 2 state-of-the-art algorithms: Many-objective Particle Swarm Optimization (MaOPSO) \cite{figueiredo2016many} and NSGA-III. We choose both algorithms due to: (i) wmoFSS incorporates many aspects of NSGA-III, such as the use of reference points and clustering the population, (ii) MaOPSO is a swarm-intelligence technique very different from wmoFSS which uses an external archive to guide the search. Results used for comparison for both algorithms were provided by MaOPSO authors and are the same used in \cite{figueiredo2016many}. 

Before comparing wmoFSS with the state-of-the-art algorithms, we performed a parameter selection step to choose the values of $step_{ind}$, $step_{vol}$ and number of fishes and also the clustering and normalization methods. In this preliminary step, we adopt a methodology inspired on the Factorial Analysis method \cite{montgomery2008design}. After this step, the results of median, maximum and minimum values of IGD for the $m$-objective DTLZ1, 2, 3 and 4 problems with $m=\{3, 5, 10\}$ obtained for wmoFSS are compared with the NSGA-III algorithm \cite{deb2014evolutionary} and the MaOPSO \cite{figueiredo2016many}.
\subsection{Results}
We show in Table \ref{FinalResults} the results obtained within the experiments involving wmoFSS, NSGA-III and MaOPSO.
\pagebreak
\begin{table}[htbp]
	\centering
	\caption{Results of median, maximum and minimum IGD obtained on the DTLZ1, 2, 3 and 4 test problems for $m=\{3, 5, 10\}$}
	\label{FinalResults}
	\begin{tabular}{c|c|ccc}
		\hline
		&                                          & wmoFSS                       & NSGA-III                     & MaOPSO                       \\ \hline
		\multirow{9}{*}{DTLZ1}                      & \multirow{3}{*}{3}                       & 3.63E-02                     & 1.51E-03                     & 6.98E-04                     \\
		&                                          & 7.27E-02                     & 1.74E-03                     & 6.99E-04                     \\
		&                                          & 1.66E-02                     & 1.37E-03                     & 6.98E-04                     \\ \cline{2-5} 
		& \multirow{3}{*}{5}                       & 1.85E-02                     & 1.59E-03                     & 8.25E-04                     \\
		&                                          & 2.27E-02                     & 1.88E-03                     & 8.25E-04                     \\
		&                                          & 9.78E-03                     & 1.50E-03                     & 8.24E-04                     \\ \cline{2-5} 
		& \multirow{3}{*}{10}                      & 1.22E-02                     & 1.42E-03                     & 1.43E-03                     \\
		&                                          & 2.32E-02                     & 2.62E-03                     & 1.44E-03                     \\
		&                                          & 8.21E-03                     & 1.38E-03                     & 1.42E-03                     \\ \hline
		\multirow{9}{*}{DTLZ2}                      & \multirow{3}{*}{3}                       & 4.44E-03                     & 3.77E-03                     & 2.27E-03                     \\
		&                                          & 4.67E-03                     & 4.11E-03                     & 2.27E-03                     \\
		&                                          & 4.24E-03                     & 3.55E-03                     & 2.27E-03                     \\ \cline{2-5} 
		& \multirow{3}{*}{5}                       & 4.71E-03                     & 1.45E-02                     & 2.94E-03                     \\
		&                                          & 4.80E-03                     & 1.53E-02                     & 2.94E-03                     \\
		&                                          & 4.62E-03                     & 1.10E-02                     & 2.94E-03                     \\ \cline{2-5} 
		& \multirow{3}{*}{10}                      & 6.07E-03                     & 1.23E-02                     & 4.95E-03                     \\
		&                                          & 6.13E-03                     & 1.25E-02                     & 4.97E-03                     \\
		&                                          & 5.97E-03                     & 9.93E-03                     & 4.94E-03                     \\ \hline
		\multicolumn{1}{l|}{\multirow{9}{*}{DTLZ3}} & \multicolumn{1}{l|}{\multirow{3}{*}{3}}  & \multicolumn{1}{l}{1.04E+00} & \multicolumn{1}{l}{3.89E-03} & \multicolumn{1}{l}{2.27E-03} \\
		\multicolumn{1}{l|}{}                       & \multicolumn{1}{l|}{}                    & \multicolumn{1}{l}{1.41E+00} & \multicolumn{1}{l}{4.36E-03} & \multicolumn{1}{l}{2.67E-01} \\
		\multicolumn{1}{l|}{}                       & \multicolumn{1}{l|}{}                    & \multicolumn{1}{l}{5.29E-01} & \multicolumn{1}{l}{3.52E-03} & \multicolumn{1}{l}{2.27E-03} \\ \cline{2-5} 
		\multicolumn{1}{l|}{}                       & \multicolumn{1}{l|}{\multirow{3}{*}{5}}  & \multicolumn{1}{l}{4.67E-01} & \multicolumn{1}{l}{1.47E-02} & \multicolumn{1}{l}{2.94E-03} \\
		\multicolumn{1}{l|}{}                       & \multicolumn{1}{l|}{}                    & \multicolumn{1}{l}{5.72E-01} & \multicolumn{1}{l}{1.91E-02} & \multicolumn{1}{l}{2.95E-03} \\
		\multicolumn{1}{l|}{}                       & \multicolumn{1}{l|}{}                    & \multicolumn{1}{l}{3.12E-01} & \multicolumn{1}{l}{4.88E-03} & \multicolumn{1}{l}{2.94E-03} \\ \cline{2-5} 
		\multicolumn{1}{l|}{}                       & \multicolumn{1}{l|}{\multirow{3}{*}{10}} & \multicolumn{1}{l}{1.69E-01} & \multicolumn{1}{l}{1.20E-02} & \multicolumn{1}{l}{4.95E-03} \\
		\multicolumn{1}{l|}{}                       & \multicolumn{1}{l|}{}                    & \multicolumn{1}{l}{2.66E-01} & \multicolumn{1}{l}{1.36E-02} & \multicolumn{1}{l}{7.54E-03} \\
		\multicolumn{1}{l|}{}                       & \multicolumn{1}{l|}{}                    & \multicolumn{1}{l}{8.83E-02} & \multicolumn{1}{l}{6.45E-03} & \multicolumn{1}{l}{4.93E-03} \\ \hline
		\multicolumn{1}{l|}{\multirow{9}{*}{DTLZ4}} & \multicolumn{1}{l|}{\multirow{3}{*}{3}}  & \multicolumn{1}{l}{8.21E-03} & \multicolumn{1}{l}{3.80E-03} & \multicolumn{1}{l}{2.45E-03} \\
		\multicolumn{1}{l|}{}                       & \multicolumn{1}{l|}{}                    & \multicolumn{1}{l}{9.29E-03} & \multicolumn{1}{l}{4.14E-03} & \multicolumn{1}{l}{2.59E-03} \\
		\multicolumn{1}{l|}{}                       & \multicolumn{1}{l|}{}                    & \multicolumn{1}{l}{7.54E-03} & \multicolumn{1}{l}{3.60E-03} & \multicolumn{1}{l}{2.36E-03} \\ \cline{2-5} 
		\multicolumn{1}{l|}{}                       & \multicolumn{1}{l|}{\multirow{3}{*}{5}}  & \multicolumn{1}{l}{6.15E-03} & \multicolumn{1}{l}{5.00E-03} & \multicolumn{1}{l}{3.88E-03} \\
		\multicolumn{1}{l|}{}                       & \multicolumn{1}{l|}{}                    & \multicolumn{1}{l}{6.58E-03} & \multicolumn{1}{l}{5.24E-03} & \multicolumn{1}{l}{4.09E-03} \\
		\multicolumn{1}{l|}{}                       & \multicolumn{1}{l|}{}                    & \multicolumn{1}{l}{5.86E-03} & \multicolumn{1}{l}{4.83E-03} & \multicolumn{1}{l}{3.51E-03} \\ \cline{2-5} 
		\multicolumn{1}{l|}{}                       & \multicolumn{1}{l|}{\multirow{3}{*}{10}} & \multicolumn{1}{l}{6.33E-03} & \multicolumn{1}{l}{5.10E-03} & \multicolumn{1}{l}{4.92E-03} \\
		\multicolumn{1}{l|}{}                       & \multicolumn{1}{l|}{}                    & \multicolumn{1}{l}{6.50E-03} & \multicolumn{1}{l}{5.20E-03} & \multicolumn{1}{l}{4.95E-03} \\
		\multicolumn{1}{l|}{}                       & \multicolumn{1}{l|}{}                    & \multicolumn{1}{l}{6.22E-03} & \multicolumn{1}{l}{5.02E-03} & \multicolumn{1}{l}{4.90E-03} \\ \hline
	\end{tabular}
\end{table}

From Table \ref{FinalResults}, one can notice that, besides the multi-modal problems, wmoFSS shows competitive performance in comparison with NSGA-III and MaOPSO. Regarding NSGA-III, wmoFSS achieved results of the same order for the 3-objective DTLZ2 and outperformed this algorithm on the 5 and 10-objective DTLZ2. For the DTLZ4 problem, wmoFSS did not outperformed NSGA-III in any of the cases considered, but achieved median, maximum and minimum IGD of the same order as NSGA-III. In comparison with MaOPSO, wmoFSS did not achieved better results for either DTLZ2 or DTLZ4, but the median, maximum and minimum IGD obtained have the same order of magnitude. From these results, it is also possible to see that wmoFSS did not achieve good results for the DTLZ1 and DTLZ3 problems. This problems are multi-modal instances of the DTLZ test suite and have multiple local PFs. We attribute this difficulty of the proposed to technique to the smaller capability of a greedy search process does not get trapped into local optima. 

\begin{figure}[h]
\centering
\includegraphics[width=0.4\textwidth]{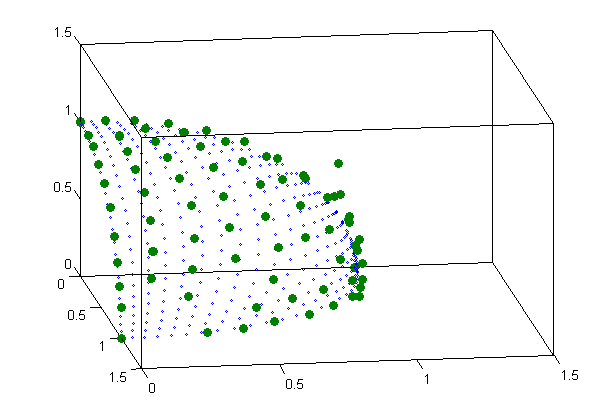}
\caption{Scatter plot of the solutions returned by wmoFSS on the 3-objective DTLZ2 problem}
\label{wmoFSSfig}
\end{figure}

\begin{figure}[h]
	\centering
	\includegraphics[width=0.4\textwidth]{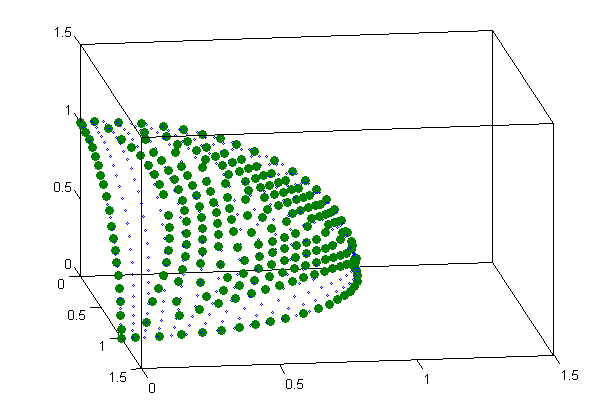}
	\caption{Scatter plot of the solutions returned by MaOPSO on the 3-objective DTLZ2 problem}
    \label{MaOPSO}
\end{figure}

\begin{figure}[h]
	\centering
	\includegraphics[width=0.4\textwidth]{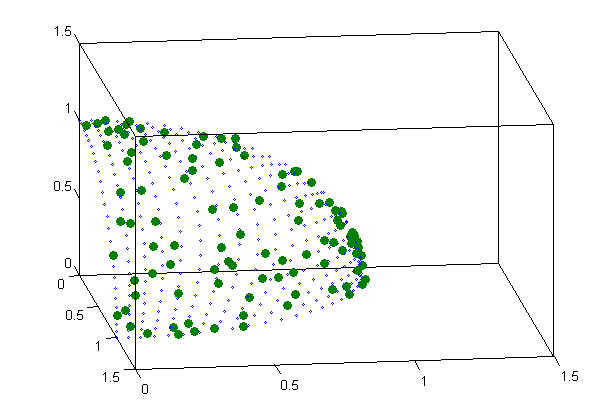}
	\caption{Scatter plot of the solutions returned by NSGA-III on the 3-objective DTLZ2 problem}
    \label{NSGAIII}
\end{figure}

To provide a graphical analysis of results, we show in Figures \ref{wmoFSSfig}, \ref{MaOPSO} and \ref{NSGAIII} scatter plots of the set of solutions returned by the compared techniques for the 3-objective DTLZ2 problem. In both cases the best IGD results were selected. A random sample of the true PF is represented by the smaller dots and the set returned by each algorithm is represented by the bigger dots. 

One can notice that, besides the fact that wmoFSS has worse IGD in this instance, the set of solutions returned by wmoFSS is able to cover some regions of the PF that the MaOPSO was not able to. Regarding the NSGA-III algorithm, Figure \ref{NSGAIII} shows that this algorithm is able to multiple solutions on the true PF with good diversity, but the returned set is not uniformly spread as the solution returned by wmoFSS.

\section{wmoFSS Solving Multi-modal DTLZ problems}\label{SBX}
As we showed in the last sections, wmoFSS algorithm is not able to achieve good results for the considered multi-modal instances of DTLZ test suite. Thus, in this section, we propose some modifications in the algorithm in order to increase its search effectiveness and then check whether the proposed changes is able to help wmoFSS to achieve better results on the DTLZ1 and DTLZ3 problems.

Within the development of this work, we notice that there was room for improvement on FSS's Individual component of the movement. This affirmative can be justified by the fact this component is calculated through a random search with depth limited by the parameter $step_{ind}$. Therefore, when a fish is moved by this component, it goes towards a direction found randomly and the algorithm does not take advantage to the information of the search process it already has at hand. Taking this fact into account, we decided to use the Simulated Binary Crossover (SBX) \cite{deb1995simulated} based approach in order to determine the direction a fish will move within the Individual component of the movement, giving rise to the wmoFSS-SBX algorithm. The SBX operator uses as parents two individuals from the current population to create two new solutions, as the regular crossover operators. However, the main difference of SBX is that it allows the creation of solutions near their parents. Hence, the goal is to use SBX to crossover a fish and its leader to create a new position and use it to generate movement directions during the computation of the Individual component.

Considering a fish $\textbf{x}(t)$ with leader $\textbf{x}_l(t)$, to calculate its candidate position of the Individual component of the movement $\textbf{x}(t+1)$, firstly, it is necessary to create a new position in the search space with the SBX taking as parents $\textbf{x}$ and $\textbf{x}_l$. The $i$-th component of the generated position $\textbf{y}(t)$ is given by:

\begin{eqnarray}
\begin{small}
\begin{cases}
0.5\left[\left(x_i(t) + x_{l,i}(t) \right) - \overline{\beta}|x_i(t) - x_{l,i}(t)|\right], \text{if} \, v \leq 0.5  \\
0.5\left[\left(x_i(t) + x_{l,i}(t) \right) + \overline{\beta}|x_i(t) - x_{l,i}(t)|\right], \text{if} \, v > 0.5 \end{cases}
\end{small}
\end{eqnarray}
with
\begin{equation}\label{etaSBX}
\overline{\beta} = \begin{cases}
(\alpha u)^{\frac{1}{\eta_c +1}}, \quad \text{if} \quad u \leq \frac{1}{\alpha}  \\
(\frac{1}{2-\alpha u})^{\frac{1}{\eta_c +1}}, \quad \text{otherwise} \end{cases}
\end{equation}
where $u$ and $v$ are uniformly distributed real random numbers, $\eta_c$ is a parameter named Distribution Index and the definition of $\alpha$ can be found at \cite{deb1995simulated}. 
After computing $\textbf{y}(t)$, the candidate position is calculated according to:
\begin{equation}\label{SBXIndMov}
\textbf{x}(t+1) = \textbf{x}(t) + step_{ind}(t)\frac{(\textbf{y}(t) - \textbf{x}(t))}{||\textbf{y}(t) - \textbf{x}(t)||}.
\end{equation}
From Equation \ref{SBXIndMov} one can see that in wmoFSS-SBX fishes move to positions determined by a SBX instead of random directions. After the calculation of $\textbf{x}(t+1)$, the algorithm follows the same steps as wmoFSS. Thus, the aggregated weight of $\textbf{x}(t+1)$ is computed and compared with the aggregated weight of $\textbf{x}(t)$. In wmoFSS-SBX we decided to use the SAR mode 0, that is, there is no worsening allowance and the fish moves to $\textbf{x}(t+1)$ only if its aggregated weight is better. The Feeding, Clustering, leaders definition and collective operators remains the same as in wmoFSS.

In order to evaluate wmoFSS-SBX we utilized the same $step_{ind}$, $step_{vol}$, normalization procedure and clustering method selected for wmoFSS. As in the work of Deb and Agrawal \cite{deb1995simulated} we decided to use $\eta_c = 1$, aiming to obtain solutions close to the fathers, but not as close as the common use in literature, where authors choose higher values for $\eta_c$, such as 20 \cite{zhang2007moea} or 30 \cite{deb2014evolutionary}, \cite{li2015evolutionary}, \cite{yuan2016new}, \cite{xiangvector}. This decision was taken in order to provide to a fish a more diverse set of possibilities for its movement and improve its chance of going to a better position. 

Another important aspect of the wmoFSS-SBX is its greater dependence on the leader of the cluster. In the wmoFSS, the role of the leader of a cluster is just not to be affected by the collective operators, which represents a type of elitism. On the other hand, in wmoFSS-SBX the leader plays a key role during the search process because it has a great impact on the generation of new solutions. Hence, it is fundamental for wmoFSS-SBX a cluster be able to improve its leader during the search. In order to do so, we increase the number of fishes per cluster by decreasing the number of reference points and also increasing the school size to 1000 fishes. 


\subsection{Study of the influence of collective operators on wmoFSS-SBX}
Within preliminary tests on wmoFSS-SBX, we notice that, in some cases, the Collective-instinctive component of the movement and the Collective-volitive component of the movement seemed to decrease the algorithm performance, mostly the Collective-instinctive component. Hence, this hypothesis was verified according to the following experiments. First, three versions of wmoFSS-SBX were defined: (A): only Individual and Collective-volitive components; (B): only Individual component; (C): all components. Then, two values of the $\theta$ parameter were used. This aspect was studied in these experiments due to the fact in wmoFSS-SBX there are more fishes per cluster, which helps the algorithm to ensure that each subproblem will be optimally solved, increasing the diversity of the results. Hence, we want to observe the behavior of wmoFSS-SBX when we decrease $\theta$ from 5 to 1, emphasizing convergence in spite of diversity. 

The experimental setup employed was: 20 runs and 10000 iterations, consider as preliminary performance measures mean and standard deviation of IGD, and analyze statistical significance of results using the Kruskal-Wallis test with $95 \% $ of confidence. Results of mean and standard deviation of the IGD obtained by the three versions are presented in Table \ref{SBXABC}. Results of Kruskal-Wallis test are also shown in Table \ref{SBXABC}, where $+$ indicates that a certain value of $\theta$ was significant better, $-$ indicates the opposite, and $=$ indicates that IGD values achieved with both $\theta$ are indifferent.
\begin{table}[h]
	\centering
	\caption{Results of mean and standard deviation for IGD obtained by wmoFSS-SBX on the 5-objective DTLZ1 and DTLZ3 problems}
	\label{SBXABC}
	\begin{tabular}{c|c|ccc}
		\hline
		&                    & $\theta$ & Mean     & SD       \\ \hline
		\multirow{6}{*}{DTLZ1} & \multirow{2}{*}{A} & 1     & \textbf{3.25E-03$^+$} & \textbf{2.57E-05$^+$} \\
		&                    & 5     & 3.55E-03 & 2.42E-04 \\ \cline{2-5} 
		& \multirow{2}{*}{B} & 1     & \textbf{3.27E-03$^+$} & \textbf{3.87E-05$^+$} \\
		&                    & 5     & 3.73E-03 & 3.00E-04 \\ \cline{2-5} 
		& \multirow{2}{*}{C} & 1     & \textbf{7.11E-03$^+$} & \textbf{2.36E-03$^+$} \\
		&                    & 5     & 1.36E-02 & 6.28E-03 \\ \hline
		\multirow{6}{*}{DTLZ3} & \multirow{2}{*}{A} & 1     & \textbf{1.81E-01$^+$} & \textbf{2.09E-01$^+$} \\
		&                    & 5     & 2.50E-01 & 2.76E-01 \\ \cline{2-5} 
		& \multirow{2}{*}{B} & 1     & \textbf{2.33E-02$^=$} & \textbf{5.19E-03$^=$} \\
		&                    & 5     & 2.53E-02 & 7.12E-03 \\ \cline{2-5} 
		& \multirow{2}{*}{C} & 1     & 2.41E+00$^-$ & \textbf{2.77E-01$^-$} \\
		&                    & 5     & \textbf{1.43E+00} & 5.69E-01 \\ \hline
	\end{tabular}
\end{table}
From Table \ref{SBXABC}, one can notice that, for the DTLZ1 problem, $\theta=1$ provided statistically significant better IGD values. On the other hand, the same is not observed for DTLZ3. Version A achieved statistically significant better results on DTLZ3 with $\theta=1$, version C with $\theta=5$ and, for version B, the use of both values of $\theta$ provided statistical indifferent results.

In order to define which is the version of wmoFSS-SBX more appropriated to solve DTLZ1 and DTLZ3, versions A, B and C are compared using the best value of $\theta$ found to solve each problem. For version A, $\theta=1$ for both DTLZ1 and DTLZ3. For version 3, $\theta=1$ for DTLZ1 and we also selected $\theta=1$ for DTLZ3, as both values are indifferent. For version C, $\theta=1$ for DTLZ1 and $\theta=5$ for DTLZ3. Results of the Kruskal-Wallis test on IGD values with $95 \%$ of confidence are shown in Table \ref{ABCKW}.

\begin{table}[htbp]
	\centering
	\caption{Results of Kruskal-Wallis tests on the IGD values for versions A, B and C of wmoFSS-SBX with the best value of $\theta$}
	\label{ABCKW}
	\begin{tabular}{c|c|ccc}
		\hline
		&   & A   & B   & C   \\ \hline
		\multirow{3}{*}{DTLZ1} & A &     & $=$ & $+$ \\
		& B & $=$ &     & $+$ \\
		& C & $-$ & $-$ &     \\ \hline
		\multirow{3}{*}{DTLZ3} & A &     & $+$ & $+$ \\
		& B & $-$ &     & $+$ \\
		& C & $-$ & $-$ &     \\ \hline
	\end{tabular}
\end{table}

From Table \ref{ABCKW}, one can notice that version B with $\theta=1$ is the best choice among the versions tested to solve DTLZ3. For DTLZ1, the performances of versions A and B are statistically indifferent, which leads us to conclude that the presence of the Collective-volitive component has no effect in the search for the considered case. We decided to tackle DTLZ1 with version B due to the fact it demands less computational effort, as it does not necessitate to calculate the Collective-volitive component.

\subsection{Comparison between wmoFSS and wmoFSS-SBX}
Now that a version of wmoFSS-SBX that better solve DTLZ1 and DTLZ3 was selected, we are able to compare it with wmoFSS. As we did in the comparison of wmoFSS with NSGA-III and MaOPSO, both algorithms solved the $\{3, 5, 10\}$- objective DTLZ1, 2, 3 and 4 problems and the obtained median, maximum and minimum IGD values among the 20 runs are taken into account in the analysis. Table \ref{OrigAndSBX} shows the results.

\begin{figure}[h]
	\centering
	\includegraphics[width=0.4\textwidth]{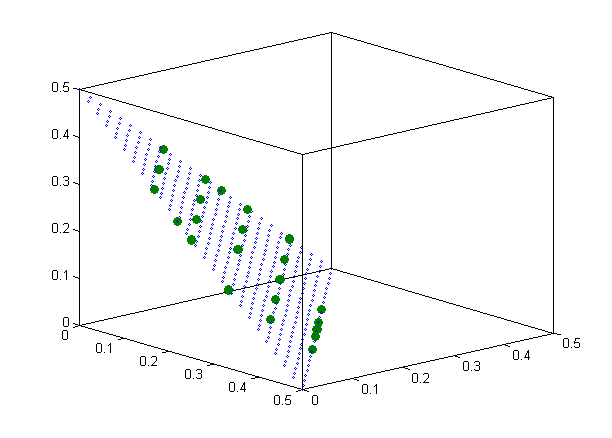}
	\caption{Scatter plot of the solutions returned by wmoFSS-SBX for the 3-objective DTLZ1.}
    \label{SBXDTLZ1}
\end{figure}

\begin{figure}[h]
	\centering
	\includegraphics[width=0.4\textwidth]{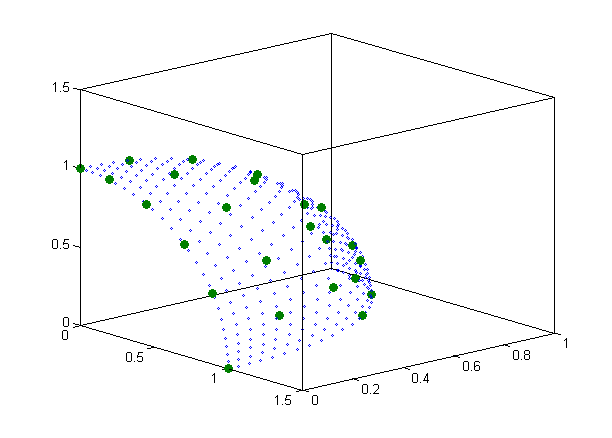}
	\caption{Scatter plot of the solutions returned by wmoFSS-SBX for the 3-objective DTLZ2.}
    \label{SBXDTLZ2}
\end{figure}

\begin{figure}[h]
	\centering
	\includegraphics[width=0.4\textwidth]{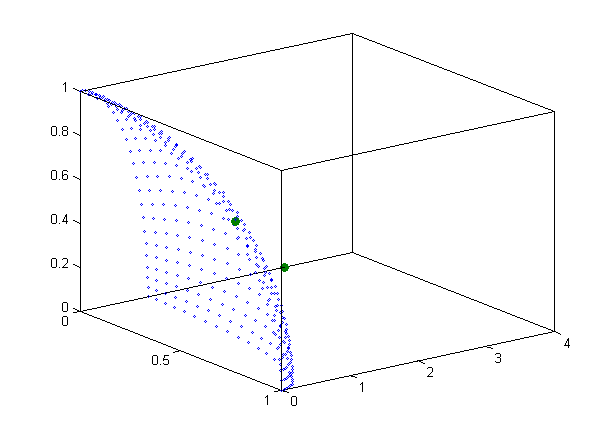}
	\caption{Scatter plot of the solutions returned by wmoFSS-SBX for the 3-objective DTLZ3.}
    \label{SBXDTLZ3}
\end{figure}

\begin{figure}[h]
	\centering
	\includegraphics[width=0.4\textwidth]{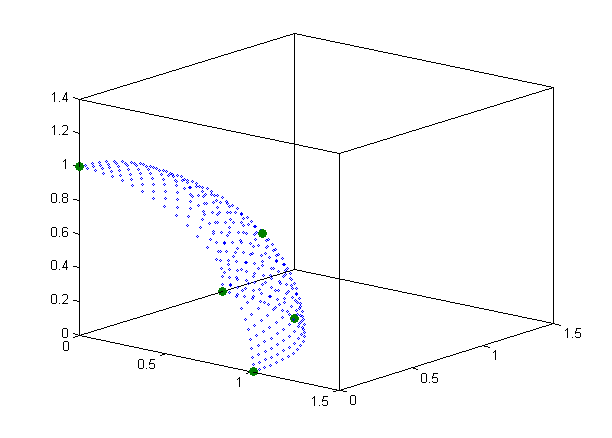}
	\caption{Scatter plot of the solutions returned by wmoFSS-SBX for the 3-objective DTLZ4.}
    \label{SBXDTLZ4}
\end{figure}

\begin{table}[h]
\centering
	\caption{Results of median, max. and min. IGD obtained by wmoFSS and wmoFSS-SBX on DTLZ1, 2, 3 and 4 for $m=\{3, 5, 10\}$.}
	\label{OrigAndSBX}
\begin{tabular}{cccc}
\hline
                                            &                                          & wmoFSS   & wmoFSS-SBX \\ \hline
\multicolumn{1}{c|}{\multirow{9}{*}{DTLZ1}} & \multicolumn{1}{c|}{\multirow{3}{*}{3}}  & 3.63E-02 & 4.00E-03   \\
\multicolumn{1}{c|}{}                       & \multicolumn{1}{c|}{}                    & 7.27E-02 & 5.16E-03   \\
\multicolumn{1}{c|}{}                       & \multicolumn{1}{c|}{}                    & 1.66E-02 & 3.48E-03   \\ \cline{2-4} 
\multicolumn{1}{c|}{}                       & \multicolumn{1}{c|}{\multirow{3}{*}{5}}  & 1.85E-02 & 3.28E-03   \\
\multicolumn{1}{c|}{}                       & \multicolumn{1}{c|}{}                    & 2.27E-02 & 3.37E-03   \\
\multicolumn{1}{c|}{}                       & \multicolumn{1}{c|}{}                    & 9.78E-03 & 3.21E-03   \\ \cline{2-4} 
\multicolumn{1}{c|}{}                       & \multicolumn{1}{c|}{\multirow{3}{*}{10}} & 1.22E-02 & 2.41E-03   \\
\multicolumn{1}{c|}{}                       & \multicolumn{1}{c|}{}                    & 2.32E-02 & 2.49E-03   \\
\multicolumn{1}{c|}{}                       & \multicolumn{1}{c|}{}                    & 8.21E-03 & 2.35E-03   \\ \hline
\multicolumn{1}{c|}{\multirow{9}{*}{DTLZ2}} & \multicolumn{1}{c|}{\multirow{3}{*}{3}}  & 4.44E-03 & 8.44E-03   \\
\multicolumn{1}{c|}{}                       & \multicolumn{1}{c|}{}                    & 4.67E-03 & 9.34E-03   \\
\multicolumn{1}{c|}{}                       & \multicolumn{1}{c|}{}                    & 4.24E-03 & 6.79E-03   \\ \cline{2-4} 
\multicolumn{1}{c|}{}                       & \multicolumn{1}{c|}{\multirow{3}{*}{5}}  & 4.71E-03 & 1.06E-02   \\
\multicolumn{1}{c|}{}                       & \multicolumn{1}{c|}{}                    & 4.80E-03 & 1.23E-02   \\
\multicolumn{1}{c|}{}                       & \multicolumn{1}{c|}{}                    & 4.62E-03 & 8.70E-03   \\ \cline{2-4} 
\multicolumn{1}{c|}{}                       & \multicolumn{1}{c|}{\multirow{3}{*}{10}} & 6.07E-03 & 9.95E-03   \\
\multicolumn{1}{c|}{}                       & \multicolumn{1}{c|}{}                    & 6.13E-03 & 1.05E-02   \\
\multicolumn{1}{c|}{}                       & \multicolumn{1}{c|}{}                    & 5.97E-03 & 8.94E-03   \\ \hline
\multicolumn{1}{c|}{\multirow{9}{*}{DTLZ3}} & \multicolumn{1}{c|}{\multirow{3}{*}{3}}  & 1.04E+00 & 6.79E-02   \\
\multicolumn{1}{c|}{}                       & \multicolumn{1}{c|}{}                    & 1.41E+00 & 1.42E-01   \\
\multicolumn{1}{c|}{}                       & \multicolumn{1}{c|}{}                    & 5.29E-01 & 3.19E-02   \\ \cline{2-4} 
\multicolumn{1}{c|}{}                       & \multicolumn{1}{c|}{\multirow{3}{*}{5}}  & 2.18E-02 & 2.18E-02   \\
\multicolumn{1}{c|}{}                       & \multicolumn{1}{c|}{}                    & 5.72E-01 & 3.93E-02   \\
\multicolumn{1}{c|}{}                       & \multicolumn{1}{c|}{}                    & 3.12E-01 & 1.63E-02   \\ \cline{2-4} 
\multicolumn{1}{c|}{}                       & \multicolumn{1}{c|}{\multirow{3}{*}{10}} & 1.69E-01 & 2.42E-02   \\
\multicolumn{1}{c|}{}                       & \multicolumn{1}{c|}{}                    & 2.66E-01 & 3.57E-02   \\
\multicolumn{1}{c|}{}                       & \multicolumn{1}{c|}{}                    & 8.83E-02 & 1.53E-02   \\ \hline
\multicolumn{1}{c|}{\multirow{9}{*}{DTLZ4}} & \multicolumn{1}{l|}{\multirow{3}{*}{3}}  & 8.21E-03 & 2.33E-02   \\
\multicolumn{1}{c|}{}                       & \multicolumn{1}{l|}{}                    & 9.29E-03 & 2.76E-02   \\
\multicolumn{1}{c|}{}                       & \multicolumn{1}{l|}{}                    & 7.54E-03 & 1.37E-02   \\ \cline{2-4} 
\multicolumn{1}{c|}{}                       & \multicolumn{1}{l|}{\multirow{3}{*}{5}}  & 6.15E-03 & 1.05E-02   \\
\multicolumn{1}{c|}{}                       & \multicolumn{1}{l|}{}                    & 6.58E-03 & 1.42E-02   \\
\multicolumn{1}{c|}{}                       & \multicolumn{1}{l|}{}                    & 5.86E-03 & 8.66E-03   \\ \cline{2-4} 
\multicolumn{1}{c|}{}                       & \multicolumn{1}{l|}{\multirow{3}{*}{10}} & 6.33E-03 & 9.31E-03   \\
\multicolumn{1}{c|}{}                       & \multicolumn{1}{l|}{}                    & 6.50E-03 & 1.03E-02   \\
\multicolumn{1}{c|}{}                       & \multicolumn{1}{l|}{}                    & 6.22E-03 & 8.26E-03   \\ \hline
\end{tabular}
\end{table}
From Table \ref{OrigAndSBX}, one can observe that the use of SBX in the Individual component and all modifications incorporated by wmoFSS-SBX, in fact, improved the results on DTLZ1 and DTLZ3. The performance on DTLZ2 and DTLZ4 was worse. To have a more detailed analysis of the results we show in Figures \ref{SBXDTLZ1}, \ref{SBXDTLZ2}, \ref{SBXDTLZ3}, and \ref{SBXDTLZ4} the scatter plot of the run for which wmoFSS-SBX obtained the best IGD on the 3-objective DTLZ1, 2, 3 and 4. The smaller dots represent the true PF and the bigger dots represent the set of solutions found by wmoFSS.   

\newpage

From Figure \ref{SBXDTLZ1} it is possible to notice that, unlike wmoFSS, wmoFSS-SBX is able to find solutions of the true PF of the 3-objective DTLZ1 problem. However, we can also see that the returned set of solutions did not cover all the PF and the extreme regions of the PF are not represented in the PF found by wmoFSS-SBX. Observing Figures \ref{SBXDTLZ2} and \ref{SBXDTLZ4} one can notice that wmoFSS-SBX is also able to find solutions on the true PF of the 3-objective DTLZ2 and DTLZ4 but with poor spread within the PF. This fact is also evidenced by the worse IGD results obtained by wmoFSS-SBX in Table \ref{OrigAndSBX}. We attribute this to the use of fewer reference points by wmoFSS-SBX, which decreases the number of sub-problems to be solved and, as consequence, the diversity of the solutions. At last, from Figure \ref{SBXDTLZ3}, we see that wmoFSS-SBX is able to find only one solution on the true PF of the 3-objective DTLZ3 problem, which is not a good result, but consists in an improvement in comparison with wmoFSS, which is not able to find any solution in the true PF of this test instance.

Summarizing, wmoFSS-SBX indeed presented a better performance on DTLZ1 and DTLZ3 and was also able to find optimal solutions on DTLZ2 and DTLZ4 problems, but with a much poor diversity in comparison with wmoFSS, mostly in the DTLZ4 problem, for which wmoFSS-SBX was able to find a small set of solutions.

\section{Conclusion} \label{Conclusion}
In this work we conceived, implemented and tested the Weight-based Many-Objective Fish School Search algorithm (wmoFSS), an adaptation of the metaheuristic Fish School Search to solve many-objective optimization problems. Our goal was to propose a competitive approach that encompasses the necessary aspects and concepts desirably the ones readily applied by state-of-the-art algorithms.

When comparing wmoFSS with state-of-the-art algorithms, we concluded that the proposed technique achieved competitive results in the uni-modal DTLZ instances, as it achieved IGD values of the same magnitude order as the other algorithms, and outperformed NSGA-III on the $\{5,10\}$-objective DTLZ2.  

To provide to the initial version of wmoFSS ways to solve multi-modal instances of DTLZ problems, we substituted the random search on FSS's Individual component of the movement by a local search guided by directions obtained with SBX crossover taking a solution and its leader as parents. The goal of this modification is to use readily available information to improve wmoFSS search effectiveness. This version was called wmoFSS-SBX. 

wmoFSS-SBX has shown to be more effective than wmoFSS to solve DTLZ1 and DTLZ3. On the other hand, it decreased the diversity of the solutions found for the DTLZ2 and DTLZ4. For DTLZ1, wmoFSS-SBX was able to find a set of solutions well spread within the true Pareto-optimal front, but the extreme regions were not covered. For DTLZ3, wmoFSS-SBX was able to find only one solutions in the true Pareto-optimal front, which is a result with very poor diversity, even so its performance is better than wmoFSS's, that was not able to find a single solution on the true Pareto-optimal front.           

In summary, wmoFSS shows a promising performance in the solution of multiple instances of DTLZ problems. Furthermore, wmoFSS-SBX seems to be a good direction in order to improve wmoFSS to solve multi-modal DTLZ instances. We point out in the next section future works to improve wmoFSS performance and to analyze its behavior in more details. As future works, we suggest the use of a non-normalized test suite, such as the WFG \cite{huband2006review}. Furthermore, Estimation of Distribution Algorithm (EDA) \cite{larranaga2012review} can be used to generate movement directions in the Individual component of the movement \cite{castro2016c}. 

\bibliographystyle{IEEEtran}
\bibliography{wmoFSSpaper}







\end{document}